\documentclass[runningheads]{llncs}
\usepackage{times}
\usepackage{latexsym}
\usepackage{graphicx}
\usepackage{amsmath}
\usepackage{amssymb}
\usepackage{subfigure}
\usepackage{multirow}
\usepackage{url}
\usepackage{CJK}
\allowdisplaybreaks[4]

\title{CJRC: A Reliable Human-Annotated Benchmark DataSet for Chinese Judicial Reading Comprehension}
\author{
Xingyi Duan\inst{1} \and Baoxin Wang\inst{1} \and Ziyue Wang\inst{1} \and Wentao Ma\inst{1} \and Yiming Cui\inst{1,2} \and Dayong Wu\inst{1} \and Shijin Wang\inst{1} \and Ting Liu\inst{2} \and Tianxiang Huo\inst{3} \and Zhen Hu\inst{3} \and Heng Wang\inst{3} \and Zhiyuan Liu\inst{4}
}
\institute{
Joint Laboratory of HIT and iFLYTEK (HFL), iFLYTEK Research, Beijing, China \and
Research Center for Social Computing and Information Retrieval (SCIR), Harbin Institute of Technology, Harbin, China \and
China Justice Big Data Institute \and
Department of Computer Science and Technology, Tsinghua University, China
\email{
\{xyduan,bxwang2,zywang27,wtma,ymcui,dywu2,sjwang3\}@iflytek.com \\
\{ymcui,tliu\}@ir.hit.edu.cn \\
\{huotianxiang,huzhen,wangheng\}@cjbdi.com \\
lzy@tsinghua.edu.cn
}
}

\date{}

\begin{document}
\begin{CJK}{UTF8}{gbsn}
\maketitle
\begin{abstract}
We present a Chinese judicial reading comprehension (CJRC) dataset which contains approximately 10K documents and almost 50K questions with answers. The documents come from judgment documents and the questions are annotated by law experts. The CJRC dataset can help researchers extract elements by reading comprehension technology. Element extraction is an important task in the legal field. However, it is difficult to predefine the element types completely due to the diversity of document types and causes of action. By contrast, machine reading comprehension technology can quickly extract elements by answering various questions from the long document. We build two strong baseline models based on BERT and BiDAF. The experimental results show that there is enough space for improvement compared to human annotators.
\end{abstract}

\section{Introduction}

Law is closely related to people's daily life. Almost every country in the world has laws, and everyone must abide by the law, thereby enjoying rights and fulfilling obligations. Tens of thousands of cases such as traffic accidents, private lending and divorce disputes occurs every day. At the same time, many judgment documents will be formed in the process of handling these cases. The judgment document is usually a summary of the entire case, involving the fact description, the court's opinion, the verdict, etc. The relatively small number of legal staff and the uneven level of judges may lead to wrong judgments. Even the judgments in similar cases can be very different sometimes. Moreover, a large number of documents make it challenging to extract information from them. Thus, it will be helpful to introduce artificial intelligence to the legal field for helping judges make better decisions and work more effectively.

\begin{figure}[t]
  \centering
  \includegraphics[height=10.0cm,width=.8\linewidth]{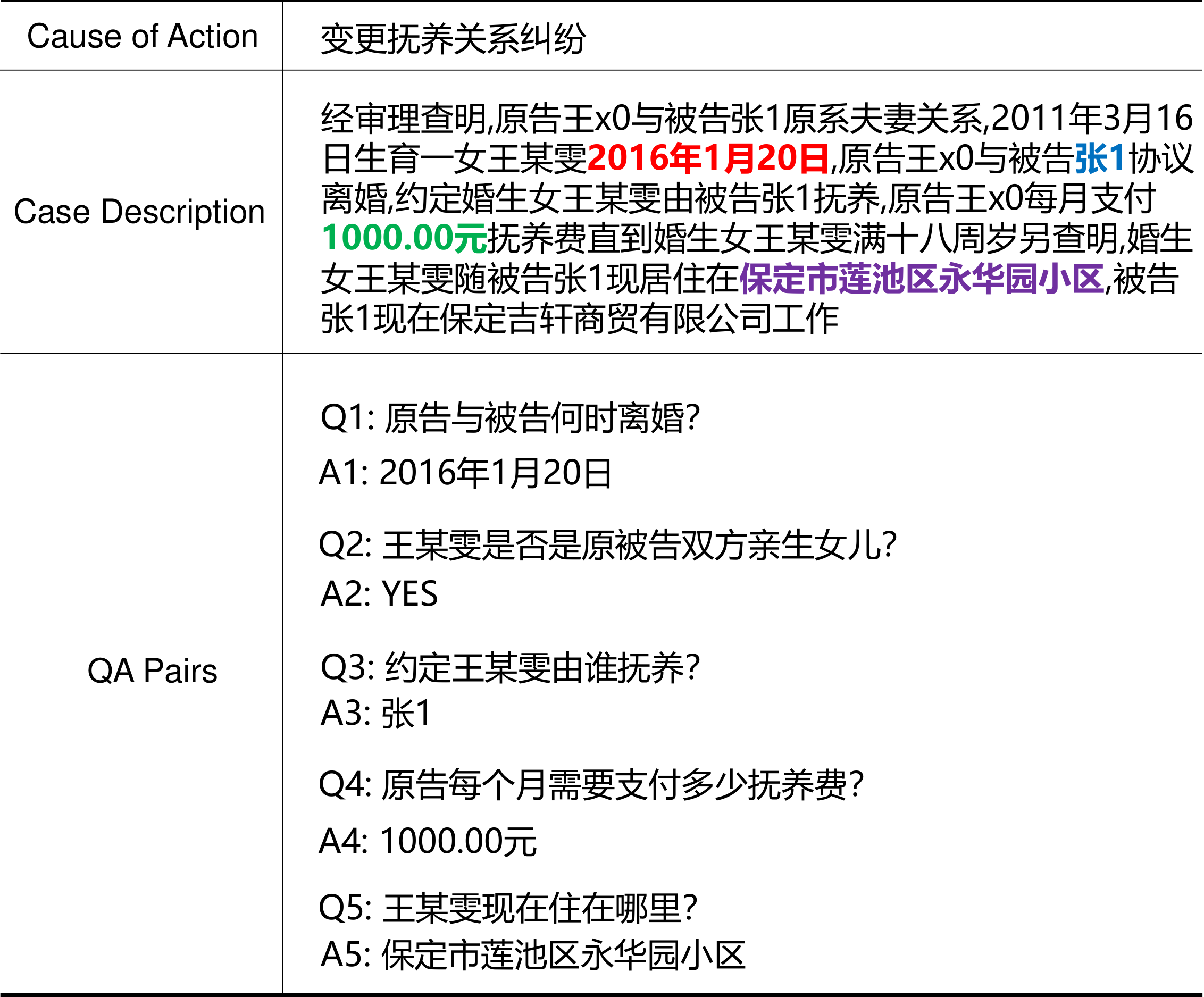}\\
  \caption{An example from the CJRC dataset. Each case contains cause of action(or called charge for criminal cases), context, and some QA pairs where yes/no and unanswerable question types are included.}\label{fig1}
\end{figure}

Currently, researchers have done amounts of work on the field of Chinese legal instruments, involving a wide variety of research aspects. Law prediction~\cite{Fawei2018,ref_tran2017applying} and charge prediction~\cite{DBLP:conf/coling/HuLT0S18,Bing2017,Zhong2018Legal} have been widely studied, especially, CAIL2018 (Chinese AI and Law challenge, 2018)~\cite{DBLP:journals/corr/abs-1807-02478,DBLP:journals/corr/abs-1810-05851} was held to predict the judgment results of legal cases including relevant law articles, charges and prison terms. Some other researches include text summarization for legal documents~\cite{Kanapala2017Text}, legal consultation~\cite{Paulo2005Ques,Ni2017Onto} and legal entity identification~\cite{Yin2018Neural}. There also exists some systems for similar cases search, legal documents correction and so on.

Information retrieval usually only returns a batch of documents in a coarse-grained manner. It still takes a lot of effort for the judges to read and extract information from document. Elements extraction often requires pre-defining element types. Different element types need to be defined for different cases or crimes. Manual definition and labeling processes are time consuming and labor intensive. These two technologies cannot cater for the fine-grained, unconstrained information extraction requirements. By contrast, reading comprehension technology can naturally extract fine-grained and unconstrained information.

In this paper, we present the first Chinese judicial reading comprehension dataset (CJRC). CJRC consists of about 10K documents which are collected from \url{http://wenshu.court.gov.cn/} published by the Supreme People’s Court of China. We mainly extract the fact description from the judgment document and ask law experts to annotate four to five question-answer pairs based on the fact. Eventually, our dataset contain around 50K questions with answers. Since some of the questions cannot be directly answered from the fact description, we have asked law experts to annotate some unanswerable and yes/no questions similar to SQuAD2.0 and CoQA datasets (Figure \ref{fig1} shows an example). In view of the fact that the civil and criminal judgment documents greatly differ in the fact description, the corresponding types of questions are not the same. This dataset covers the two types of documents and thereby covers most of the judgment documents, involving various types of charge and cause of action (in the following parts, we will use \emph{casename} to refer to civil cases and criminal charges.).

The main contribution of our work can be concluded as follows:

\begin{itemize}
  \item CJRC is the first Chinese judicial reading comprehension dataset to fill gaps in the field of legal research.
  \item Our proposed dataset includes a wide range of areas, specifically 188 causes of action and 138 criminal charges. Moreover, the research results obtained through this dataset can be widely applied, such as information retrieval and factor extraction.
  \item The performance of some powerful baselines indicates there is enough space for improvement compared to human annotators.
\end{itemize}

\begin{table}[t]
  \centering
  \setlength{\abovecaptionskip}{0.5cm}
  \begin{tabular}{|c|c|c|c|c|}
    \hline
    & Lang & \#Que & Domain & Answer Type \cr
    \hline
    \hline
    CNN/Daliy Mail & ENG & 1.4M & News & Fill in entity \cr
    RACE & ENG & 870K & English Exam &  Multi. choices \cr
    NewsQA & ENG & 100K & CNN & Span of words \cr
    SQuAD & ENG & 100K & Wiki & Span of words, Unanswerable \cr
    CoQA & ENG & 127K & Children's Sto. etc. & Span of words, yes/no, unanswerable \cr
    TriviaQA & ENG & 40K & Wiki/Web doc &  Span/substring of words \cr
    HFL-RC & CHN & 100K & Fairy/News & Fill in word \cr
    DuReader & CHN & 200K & Baidu Search/Baidu Zhidao & Manual summary \cr
    \hline
    \textbf{CJRC} & \textbf{CHN} & \textbf{50K} & \textbf{Law} & \textbf{Span of words, yes/no, unanswerable} \cr\hline
  \end{tabular}
  \caption{Comparison of CJRC with existing reading comprehension datasets}\label{tab1}
\end{table}

\section{Related Work}

\subsection{Reading Comprehension Datasets}

Machine reading comprehension (MRC) has emerged a few datasets for researches. Among these data sets, English reading comprehension datasets occupy a large proportion. Almost each of the mainstream datasets is designed to cater for demands of requiring specific scenes or domains corpus, or to solve one or more certain problems. CNN/Daliy mail~\cite{CNN2015Hill} and NewsQA~\cite{DBLP:journals/corr/TrischlerWYHSBS16} refer to news field, SQuAD 2.0~\cite{DBLP:journals/corr/abs-1806-03822} focuses on wikipedia, and RACE~\cite{DBLP:journals/corr/LaiXLYH17} concentrates on Chinese middle school students' English reading comprehension examination questions. SQuAD 2.0~\cite{DBLP:journals/corr/abs-1806-03822} mainly introduces the unanswerable questions due to the real situations that we sometimes cannot find a favourable answer according to a given context. CoQA~\cite{DBLP:journals/corr/abs-1808-07042} is a large-scale reading comprehension dataset which contains questions that depend on a conversation history. TriviaQA~\cite{DBLP:journals/corr/TrischlerWYHSBS16} and SQuAD 2.0~\cite{DBLP:journals/corr/JoshiCWZ17} pay attention to complex reasoning questions, which means that we need to jointly infer the answers via multiple sentences.

Compared with English datasets, Chinese reading comprehension datasets are quite rare. HFL-RC~\cite{Cui2016} is the first Chinese Cloze-style reading comprehension dataset, and it is collected from People Daily and Children's Fairy Tale. DuReader~\cite{DBLP:journals/corr/abs-1711-05073} is an open-domain Chinese reading comprehension dataset, and it is based on Baidu Search and Baidu Zhidao. Our dataset is the first Chinese judicial reading comprehension dataset, and contains multiple types of questions. Table~\ref{tab1} compares the above datasets with ours, mainly considering the four dimensions: language, scale of questions, domain, and answer type.

\begin{figure*}[t]
  \centering
  \setlength{\abovecaptionskip}{0.5cm}
  \includegraphics[width=1.0\linewidth,height=9.0cm]{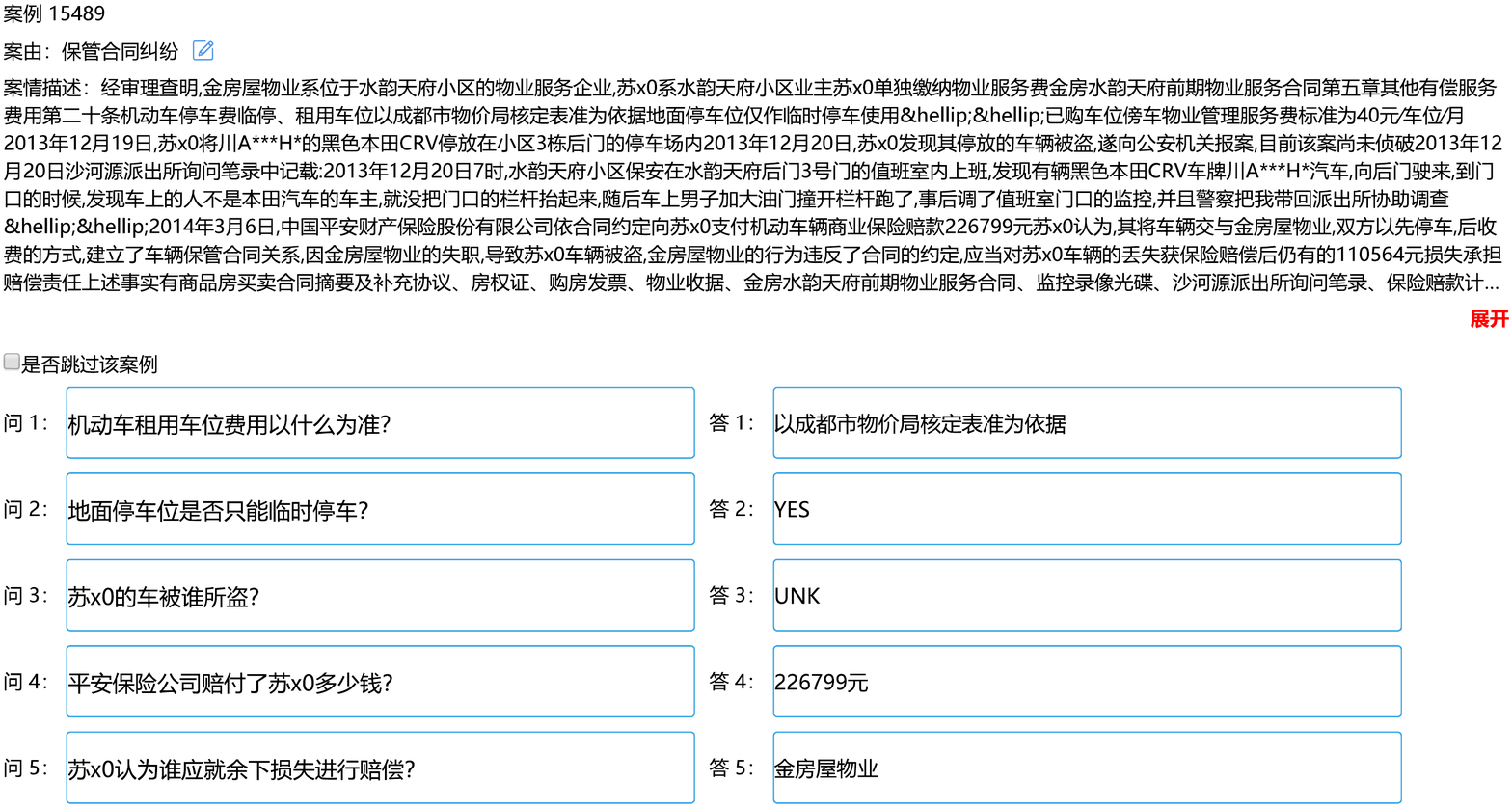}\\
  \caption{Annotate platform interface}\label{fig2}
\end{figure*}

\subsection{Reading Comprehension Models}

Cloze-style and span-extraction are two of the most widely studied tasks of MRC. Cloze-style models are usually designed as classification models to predict which word has the maximum probability. Generally, models need to encode query and document respectively into a sequence of vectors, where each vector denotes a token's representation. The next operations lead to different methods. Stanford Attentive Reader~\cite{DBLP:journals/corr/ChenBM16a} firstly obtains the query vector, and then exploits it to calculate the attention weights on all the contextual embeddings. The final document representation is computed by the weighted contextual embeddings and is used for the final classification. Some other models~\cite{DBLP:journals/corr/DhingraLCS16,DBLP:journals/corr/SukhbaatarSWF15,DBLP:journals/corr/KadlecSBK16} are similar with Stanford Attentive Reader.

Span-extraction based reading comprehension models are basically consistent in terms of the goal of calculating the start position and the end position. Some classic models are R-Net~\cite{R-NET2017}, BiDAF~\cite{DBLP:journals/corr/SeoKFH16}, BERT~\cite{DBLP:journals/corr/abs-1810-04805}, etc. BERT is a powerful pre-trained model and performs well on many NLP tasks. It is worth noting that almost all the top models on the SQuAD 2.0 leaderboard are integrated with BERT. In this paper, we use BERT and BiDAF as two strong baselines. The gap between human and BERT is 15.2\%, indicating that models still have enough room for improvement.

\section{CJRC: A New Benchmark Dataset}

Our legal documents are all collected from China Judgments Online\footnote{\url{http://wenshu.court.gov.cn/}}. We select from a batch of judgment documents, obeying the standard that the length of fact description or plaintiff's claim is not less than 150 words, where both of the two parts are extracted with regular rules. We obtain 5858 criminal documents and 5737 civil documents. We build a data annotation platform (Figure \ref{fig2}) and ask law experts to annotate QA pairs. In the following subsections, we detail how to confirm the training, development, and test sets by several steps.

\begin{figure}[htbp]
\centering
\subfigure[civil]{
\begin{minipage}[t]{0.5\linewidth}
\includegraphics[width=6cm,height=5cm]{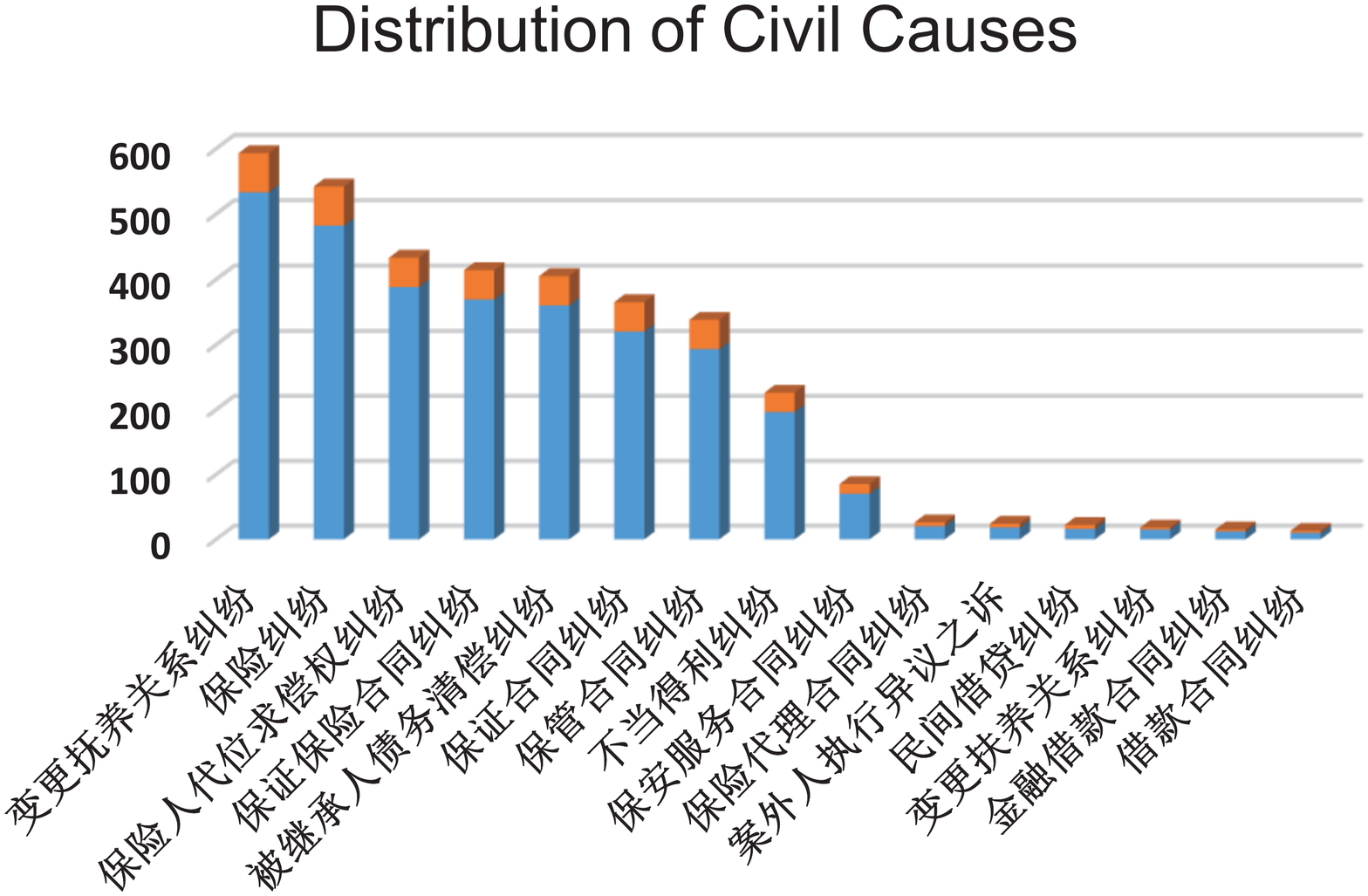}
\end{minipage}%
}%
\subfigure[criminal]{
\begin{minipage}[t]{0.5\linewidth}
\centering
\includegraphics[width=6cm,height=5cm]{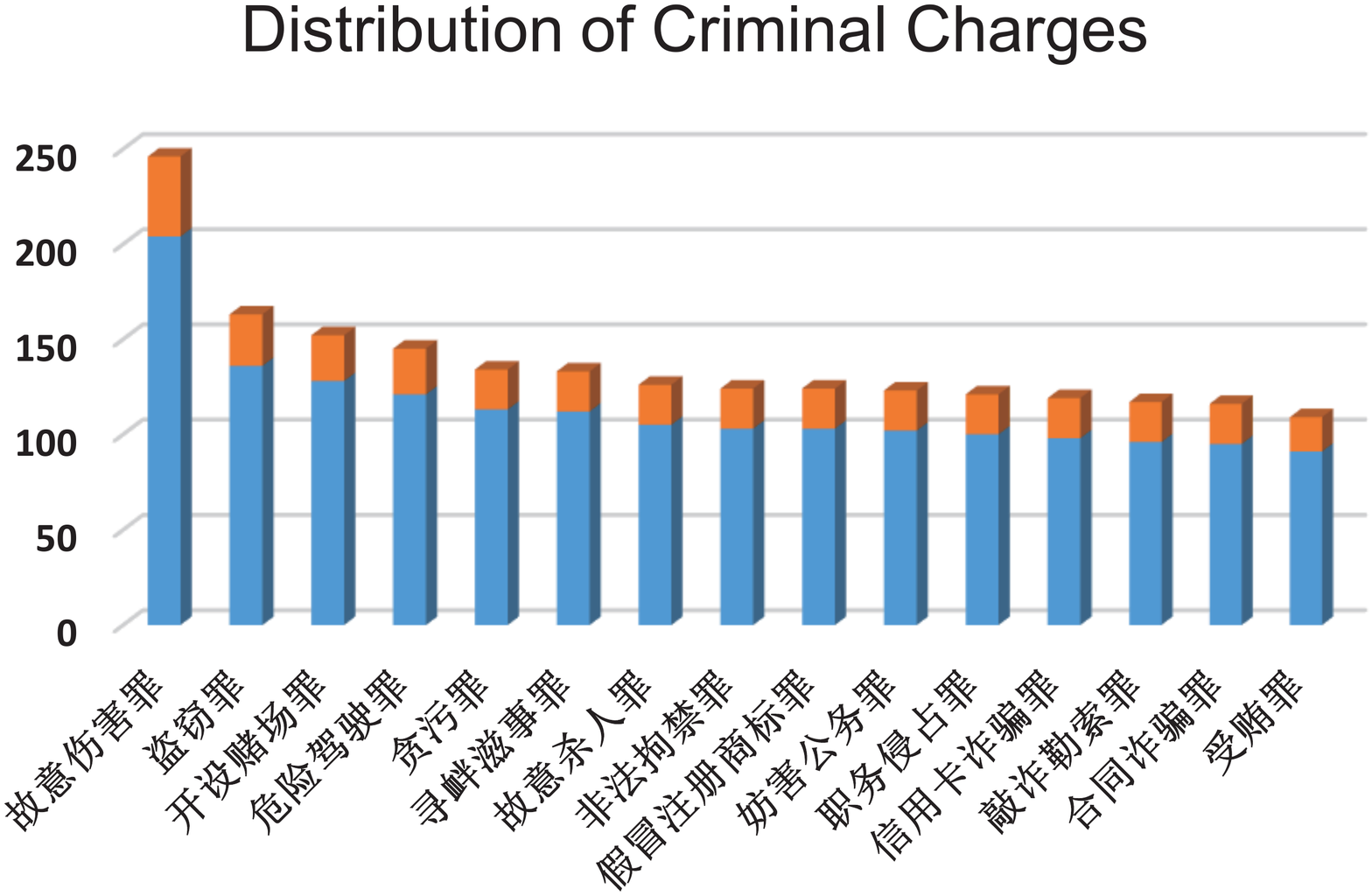}
\end{minipage}%
}%
\centering
\caption{(a) Distribution of the top 15 civil causes. (b) Distribution of the top 15 criminal charges. Blue area denotes the training set and yellow area denotes the development set.}\label{fig3}
\end{figure}

\textbf{In-domain and out-of-domain.} Referring to CoQA dataset, we divide the dataset into in-domain and out-of-domain. In-domain means that the data type of test data exists in train sets, and conversely, out-of-domain means the absence. Taking into account that \emph{casename} can be regarded as the natural segmentation attribute, we firstly determine which \emph{casenames} should be included in the training set. Then development set and test set should contain \emph{casenames} in the training set and \emph{casenames} not in the training set. Finally, we obtain totally 8000 cases for training set and 1000 cases respectively for development set and test set. For development and test set, the number of cases is the same whether it is divided by civil and criminal, or by in-domain and out-of-domain. The distribution of \emph{casenames} on the training set is shown in Figure \ref{fig3}.

\textbf{Annotate development and test sets.} After splitting the dataset, we ask annotators to annotate two extra answers for each question of each example in development and test sets. We obtain three standard answers for each question.

\textbf{Redefine the task.} Through preliminary experiments, we discovered that the distinction between in-domain and out-of-domain is not obvious. It means that performance of the model trained on training set is almost the same regarding in-domain and out-of-domain, and it is even likely that the latter works better. The possible reasons are as follows:

\begin{itemize}
  \item \emph{Casenames} inside and outside the domain are similar. In other words, the corresponding cases show some similar case issues. For example, two cases related to the contract, housing sales contract disputes and house lease contract disputes, may involve same issues such as housing agency or housing quality.
  \item Questions about time, place, etc. are more common. Moreover, due to the existence of the ``similar \emph{casenames}'' phenomenon, the corresponding questions would also be similar.
\end{itemize}

\begin{table}[t]
  \centering
  \setlength{\abovecaptionskip}{0.5cm}
  \begin{tabular}{|l|c|c|c|}
    \hline
    & Civil & Criminal & Total \\
    \hline
    \hline
    \textbf{Train} &  &  &  \\
    Total Cases & 4000 & 4000 & 8000 \\
    Total \emph{Casenames} & 126 & 53 & 179 \\
    Total Questions & 19333 & 20000 & 40000 \\
    Total Unanswerable Questions & 617 & 617 & 1901 \\
    Total Yes/No Questions & 3015 & 2093 & 5108 \\
    \hline
    \textbf{Development} &  &  &  \\
    Total Cases & 500 & 500 & 1000 \\
    Total \emph{Casenames} & 188 & 138 & 326 \\
    Total Questions & 3000 & 3000 & 6000 \\
    Total Unanswerable Questions & 685 & 561 & 1246 \\
    Total Yes/No Questions & 404 & 251 & 655 \\
    \hline
    \textbf{Test} &  &  &  \\
    Total Cases & 500 & 500 & 1000 \\
    Total \emph{Casenames} & 188 & 138 & 326 \\
    Total Questions & 3000 & 3000 & 6000 \\
    Total Unanswerable Questions & 685 & 577 & 1262 \\
    Total Yes/No Questions & 392 & 245 & 637 \\
    \hline
  \end{tabular}
  \caption{Dataset statistics of CJRC}\label{tab2}
\end{table}

However, as we all known, there are remarkable differences between civil and criminal cases. As mentioned in the module ``\textbf{In-domain and out-of-domain}'', the corpus would be divided by domain or type of cases (civil and criminal). Although we no longer consider the division of in-domain and out-of-domain, it would also make sense to train a model to perform well on both civil and criminal data.

\textbf{Adjust data distribution.} Through preliminary experiments, we also discovered that the unanswerable questions are more challenging than the other two types of questions. To increase the difficulty of the dataset, we have increased the number of unanswerable questions in development set and test set. Related experiments will be presented in the experimental section.

Via the processing of the above steps, we get the final data. Statistics of the data are shown in Table~\ref{tab2}. The subsequent experiments will be performed on the final data.

\section{Experiments}

\subsection{Evaluation Metric}

We use macro-average F1 as our evaluation metric which is consistent with the CoQA competition. For each question, $n$ F1 scores need to be calculated with $n$ standard human answers, and the maximum value is taken as its F1 score. However, in assessing human performance, each standard answer needs to be compared to $n-1$ other standard answers to calculate the F1 score. In order to compare human indicators more fairly, $n$ standard answers need to be divided into $n$ groups, where each group contains $n-1$ answers. Finally, the F1 score of each question is the average of the $n$ groups' F1. The F1 score of the entire dataset is the average of all questions' F1. The formula is as follow:

\begin{align}
  &Lg = len(gold)&\\
  &Lp = len(pred)& \\
  &Lc = InterSec(gold,pred)& \\
  &precision = \frac{Lc}{Lp}&  \\
  &recall = \frac{Lc}{Lg}& \\
  &f1(gold,pred) = \frac{2*precision*recall}{precision+recall}& \\
  &Avef1 = \frac{\sum_{i=0}^{Count_{ref}}(max(f1(gold_{\rightharpoondown i},pred))}{Count_{ref}}& \\
  &F1_{macro} = \frac{\sum_{i=1}^{N}(Avef1_i)}{N}&
\end{align}

Where $gold$ denotes standard answers, $pred$ denotes answers predicted by models, $len$ means to calculate length, $InterSec$ means to calculate the number of overlap chars. $Count_{ref}$ represents the total references, ${\rightharpoondown i}$ represents that the predicted answer is compared to all standard answers except the current one in a single group described as above.

\subsection{Baselines}

We implement and evaluate two powerful and typical model architectures: BiDAF proposed by~\cite{DBLP:journals/corr/SeoKFH16} and BERT proposed by~\cite{DBLP:journals/corr/abs-1810-04805}. Both of the two models are designed to deal with these three types of questions. These two models learn to predict the probability which is used to judge whether the question is unanswerable. In addition to the way of dealing with unanswerable questions, we concatenate [YES] and [NO] as two tokens with the context for BERT, and concatenate ``KYN'' as three chars with the context for BiDAF where `K' denoting ``Unknown'' means cannot answer the question according to the context. Taking BiDAF for example, during the prediction stage, if start index is equal to 1, then model outputs ``YES'', and if it is equal to 2, then model outputs ``NO''.

Some other implementation details: for BERT, we choose the Bert-Base Chinese pre-trained model\footnote{\url{https://github.com/google-research/bert}}, and then fine-tuning on it with our train data. It is trained on Tesla P30G24, and batch size is set to 8, max sequence length is set to 512, number of epoch is set to 2. For BiDAF, we remove the char embedding, and split string into a sequence of chars, which roles as word in English, like ``2 0 1 9 {年} 5 月 3 0 日''. We set embedding size to 300, and other parameters follow the setting in~\cite{DBLP:journals/corr/abs-1810-04805}.

\subsection{Result and Analysis}

Experimental results on test set are shown in Table~\ref{tab3}. From this table, it is obvious that BERT is 14.5$\sim$19 percentage points higher than BiDAF, and Human performance is 14.8$\sim$15.5 percentage points higher that BERT. This implies that models could be improved markedly in future research.

\begin{table}[t]
  \centering
  \setlength{\abovecaptionskip}{0.5cm}
  \begin{tabular}{|l|c|c|c|}
    \hline
    & Civil & Criminal & Overall \cr
    \hline
    \hline
    Human & 94.9 & 92.7 & 93.8 \cr\hline
    BiDAF & 61.1 & 62.7 & 61.9 \cr\hline
    BERT & 80.1 & 77.2 & 78.6 \cr\hline
  \end{tabular}
  \caption{Experimental results}\label{tab3}
\end{table}

\begin{table*}[t]
  \centering
  \setlength{\abovecaptionskip}{0.5cm}
    \begin{tabular}{|c|c|c|c|c|c|c|}
    \hline
    \multirow{2}{*}{Method}&
    \multicolumn{3}{c|}{Development}&\multicolumn{3}{c|}{Test}\cr\cline{2-7}
    &Civil&Criminal&Overall&Civil&Criminal&Overall\cr
    \hline
    \hline
    In-Domain&82.1&78.6&80.3&84.7&80.2&82.5\cr\hline
    Out-of-Domain&\textbf{82.3}&\textbf{83.9}&\textbf{83.1}&80.9&\textbf{82.9}&81.9\cr\hline
    \end{tabular}
    \caption{Experimental results of in-domain and out-of-domain on development set and test set}\label{tab4}
\end{table*}

\subsubsection{Experimental Effect of In-domain and Out-of-Domain}

In this section, we mainly explain why we no loner consider the division of in-domain and out-of-domain described in section 2. We adopts the dataset before adjusting data distribution and select BERT model to verify. Notice that we only train data belong to civil for ``Civil'', train data belong to criminal for ``Criminal'', and train all data for ``Overall''. And type of cases on development set and test set is corresponding to the training corpus. It can be seen from Table~\ref{tab4} that the F1 score of out-of-domain is even higher than that of in-domain, which obviously does not meet the expected result of setting in-domain and out-of-domain.

\subsubsection{Comparisons of Different Types of Questions}

Table~\ref{tab5} presents fine-grained results of models and humans on the development set and test set, where both of the two sets are not adjusted. We observe that humans maintain high consistency on all types of questions, especially on the ``YES'' questions. The human agreement on criminal data is lower than that on civil data. This is partly because that we firstly annotate the criminal data, and then have more experience when marking the civil data. It could result in a more consistent granularity of the selected segments on the ``Span'' questions.

Among the different question types, unanswerable questions are the hardest, and ``No'' questions are second. We analyze why the performance of unanswerable questions is the lowest, and conclude two possible causes: 1) the total number of unanswerable questions on the training set is few; 2) the unanswerable questions are more troublesome than the others.

It is easy to verify the first cause via observing the corpus. To verify the second point, we compare the unanswerable questions and the ``NO'' questions. Table~\ref{tab6} shows some comparison data of the two types of questions. The first two rows show that unanswerable questions presents a lower performance than the other on the criminal data, even though the former owns more questions. This has basically illustrated that the unanswerable questions are more hard. We have further experimented with increasing the number of unanswerable questions of civil data on the training set. The last two rows in Table~\ref{tab6} demonstrates that increasing unanswerable questions' quantity has an significant impact on performance. However, despite having a larger amount of questions for unanswerable questions, it presents a lower score than ``NO'' questions.

The above experiments could explain that the unanswerable questions are more challenging than other types of questions. To increase the difficulty of the corpus, we adjusts data distribution through controlling the number of unanswerable questions. The following section would show details about the influence of unanswerable questions.

\begin{table}[t]
  \centering
  \setlength{\abovecaptionskip}{0.5cm}
    \begin{tabular}{|c|c|c|c|c|c|c|c|c|c|}
    \hline
    \multirow{3}{*}{}&
    \multicolumn{9}{c|}{Development}\cr\cline{2-10} &
    \multicolumn{3}{c|}{Bert}&\multicolumn{3}{c|}{BiDAF}&\multicolumn{3}{c|}{Human}\cr\cline{2-10}
    &Civil&Criminal&Overall&Civil&Criminal&Overall&Civil&Criminal&Overall\cr
    \hline
    \hline
    Unanswerable&69.5&63.3&68.0&7.6&11.4&8.5&92.0&87.1&90.8\cr\hline
    YES&91.7&93.2&92.4&83.5&91.2&86.9&96.9&96.2&96.6\cr\hline
    NO&78.0&59.0&73.2&57.9&44.9&54.6&94.2&87.8&92.6\cr\hline
    Span&84.8&81.8&83.2&80.1&76.0&77.9&91.6&88.4&89.9\cr\hline
    \multirow{3}{*}{}&
    \multicolumn{9}{c|}{Test}\cr\cline{2-10} &
    \multicolumn{3}{c|}{Bert}&\multicolumn{3}{c|}{BiDAF}&\multicolumn{3}{c|}{Human}\cr\cline{2-10}
    &Civil&Criminal&Overall&Civil&Criminal&Overall&Civil&Criminal&Overall\cr
    \hline
    \hline
    Unanswerable&67.7&65.6&67.1&10.6&16.0&12.2&91.5&87.7&90.4\cr\hline
    YES&91.8&95.6&93.4&77.3&92.8&83.7&97.3&96.5&96.9\cr\hline
    NO&72.9&69.7&71.8&47.8&43.3&46.3&96.3&92.5&95.0\cr\hline
    Span&84.3&82.4&83.3&79.1&76.2&77.6&93.5&90.9&92.2\cr\hline
    \end{tabular}
    \caption{Comparisons of different types of questions.}\label{tab5}
\end{table}

\begin{table*}[t]
  \centering
  \setlength{\abovecaptionskip}{0.5cm}
    \begin{tabular}{|c|c|c|c|c|c|c|}
    \hline
    \multirow{3}{*}{}&
    \multicolumn{2}{|c|}{Number of Questions}&\multicolumn{2}{c|}{Number of Questions}&\multicolumn{2}{c|}{Performance}\cr &
    \multicolumn{2}{c|}{(Training set)}&\multicolumn{2}{c|}{(Test set)}&\multicolumn{2}{c|}{(Test set)}\cr\cline{2-7}
    &Civil&Criminal&Civil&Criminal&Civil&Criminal\cr
    \hline
    \hline
    Unanswerable&617&617&186&77&67.7&65.6\cr\hline
    NO&1058&485&134&67&72.9&69.7\cr\hline
    \hline
    Unanswerable+&1284&617&186&77&77.3&67.1\cr\hline
    NO&1058&485&134&67&81.6&71.1\cr\hline
    \end{tabular}
    \caption{Comparison data of unanswerable questions and ``NO'' questions, where unanswerable+ denotes adding extra unanswerable questions on the training set of the civil data.}\label{tab6}
\end{table*}

\subsubsection{Influence of Unanswerable Questions}

In this section, we mainly discuss the impact of the number of unanswerable questions on the difficulty of the entire dataset. \textbf{CJRC} represents that we only increase the number of unanswerable answers on the development and the test set without changes on the training set. \textbf{CJRC+Train} stands for adjusting all the datasets. \textbf{CJRC-Dev-Test} means no adjusting any of the datasets. \textbf{CJRC+Train-Dev-Test} means only increasing the number of unanswerable questions of the training set. From Table~\ref{tab7}, we can observe the following phenomenon:
\begin{itemize}
  \item Increasing the number of unanswerable questions in development and test sets can effectively increase the difficulty of the dataset. In terms of BERT, before adjustment, the gap with human indicator is 9.8\%, but after adjustment, the gap increases to 15.2\%.
  \item By comparing CJRC+Train and CJRC (or comparing CJRC+Train-Dev-Test and CJRC-Dev-Test), we can conclude that BiDAF cannot handle unanswerable questions effectively.
  \item Increasing the proportion of unanswerable questions in development and test sets is more effective in increasing the difficulty of the dataset, compared with reducing the number of unanswerable questions of the training set (get the conclusion by observing CJRC, CJRC+Train and CJRC-Dev-Test).
\end{itemize}

\begin{table}[t]
  \centering
  \setlength{\abovecaptionskip}{0.5cm}
    \begin{tabular}{|c|c|c|c|c|c|c|}
    \hline
    \multirow{3}{*}{}&
    \multicolumn{6}{c|}{Development}\cr\cline{2-7} &
    \multicolumn{3}{c|}{Bert}&\multicolumn{3}{c|}{BiDAF}\cr\cline{2-7}
    &Civil&Criminal&Overall&Civil&Criminal&Overall\cr
    \hline
    \hline
    Human(Before Adjust)&92.3&89.0&90.7&-&-&-\cr\hline
    Human(After Adjust)&93.6&90.8&92.2&-&-&-\cr\hline
    CJRC+Train&83.7&77.3&80.5&63.3&62.5&62.9\cr\hline
    CJRC-Dev-Test&84.0&81.8&82.9&73.7&75.0&74.3\cr\hline
    CJRC+Train-Dev-Test&84.8&81.7&83.3&73.8&74.9&74.4\cr\hline
    CJRC&82.0&76.4&79.2&62.8&63.1&63.0\cr\hline
    \multirow{3}{*}{}&
    \multicolumn{6}{c|}{Test}\cr\cline{2-7} &
    \multicolumn{3}{c|}{Bert}&\multicolumn{3}{c|}{BiDAF}\cr\cline{2-7}
    &Civil&Criminal&Overall&Civil&Criminal&Overall\cr
    \hline
    \hline
    Human(Before Adjust)&93.9&91.3&92.6&-&-&-\cr\hline
    Human(After Adjust)&94.9&92.7&93.8&-&-&-\cr\hline
    CJRC+Train&82.3&77.9&80.1&61.3&61.9&61.6\cr\hline
    CJRC-Dev-Test&83.2&82.5&82.8&72.2&74.6&73.4\cr\hline
    CJRC+Train-Dev-Test&84.5&82.1&83.3&72.6&74.0&73.3\cr\hline
    CJRC&80.1&77.2&78.6&61.1&62.7&61.9\cr\hline
    \end{tabular}
    \caption{Influence of unanswerable questions. Implement BERT and BiDAF on development set and test set. +Train stands for increasing the number of unanswerable questions on the training set. -Dev-Test means no adjusting the number of unanswerable questions on the development set and the test set.} \label{tab7}
\end{table}

\section{Conclusion}

In this paper, we construct a benchmark dataset named CJRC (Chinese Judicial Reading Comprehension). CJRC is the first Chinese judical reading comprehension, and could fill gaps in the field of legal research. In terms of the types of questions, it involves three types of questions, namely span-extraction, YES/NO
and unanswerable questions. In terms of the types of cases, it contains civil data and criminal data, where various of criminal charges and civil causes are included. We hope that researches on the dataset could improve the efficiency of judges' work. Integrating Machine reading comprehension with Information extraction or information retrieval would produce great practical value. We describe in detail the construction process of the dataset, which aims to prove that the dataset is reliable and valuable. Experimental results illustrate that there is still enough space for improvement on this dataset.

\section{Acknowledgements}
This work is supported by the National Key R$\&$D Program of China under Grant No.2018YFC0832103.

\bibliographystyle{splncs04}
\end{CJK}
\end{document}